\documentclass[conference]{IEEEtran}
\IEEEoverridecommandlockouts
\usepackage{cite}
\usepackage{amsmath,amssymb,amsfonts}
\usepackage{graphicx}
\usepackage{subcaption}
\usepackage{algorithm,algorithmic}
\usepackage{textcomp}
\usepackage{xcolor}
\def\BibTeX{{\rm B\kern-.05em{\sc i\kern-.025em b}\kern-.08em
    T\kern-.1667em\lower.7ex\hbox{E}\kern-.125emX}}
\begin{document}

\title{Learning Differentiable Particle Filter on the Fly
\thanks{* Jiaxi Li and Xiongjie Chen contributed equally to this work.}
}

\author{\IEEEauthorblockN{Jiaxi $\text{Li}^*$}
\IEEEauthorblockA{\textit{Computer Science Research Centre} \\
\textit{University of Surrey}\\
Guildford, United Kingdom \\
jiaxi.li@surrey.ac.uk}
\and
\IEEEauthorblockN{Xiongjie $\text{Chen}^*$}
\IEEEauthorblockA{\textit{Computer Science Research Centre} \\
\textit{University of Surrey}\\
Guildford, United Kingdom \\
xiongjie.chen@surrey.ac.uk}
\and
\IEEEauthorblockN{Yunpeng Li}
\IEEEauthorblockA{\textit{Computer Science Research Centre} \\
\textit{University of Surrey}\\
Guildford, United Kingdom \\
yunpeng.li@surrey.ac.uk}
}

\maketitle

\begin{abstract}
Differentiable particle filters are an emerging class of sequential Bayesian inference techniques that use neural networks to construct components in state space models. Existing approaches are mostly based on offline supervised training strategies. This leads to the delay of the model deployment and the obtained filters are susceptible to distribution shift of test-time data. In this paper, we propose an online learning framework for differentiable particle filters so that model parameters can be updated as data arrive. The technical constraint is that there is no known ground truth state information in the online inference setting. We address this by adopting an unsupervised loss to construct the online model updating procedure, which involves a sequence of filtering operations for online maximum likelihood-based parameter estimation. We empirically evaluate the effectiveness of the proposed method, and compare it with supervised learning methods in simulation settings including a multivariate linear Gaussian state-space model and a simulated object tracking experiment.
\end{abstract}

\begin{IEEEkeywords}
differentiable particle filters, sequential Bayesian inference, online learning.
\end{IEEEkeywords}

\section{Introduction}
Particle filters, or sequential Monte Carlo (SMC) methods, are a class of simulation-based algorithms designed for solving recursive Bayesian filtering problems in state-space models~\cite{gordon1993novel,djuric2003particle,doucet2000sequential}. Although particle filters have been successfully applied in a variety of challenging tasks, including target tracking, econometrics, and fuzzy control~\cite{qian20173d, creal2012survey, pozna2022hybrid}, the design of particle filters' components relies on practitioners' domain knowledge and often involves a trial and error approach. To address this challenge, various parameter estimation techniques have been proposed for particle filters~\cite{kantas2009overview,kantas2015particle}. However, many such methods are restricted to scenarios where the structure or the parameters of the state-space model are partially known.

For real-world applications where both the structure and the parameters of the considered state-space model are unknown, differentiable particle filters (DPFs) were proposed to construct the components of particle filters with neural networks and adaptively learn their parameters from data~\cite{corenflos2021differentiable,karkus2018particle,chen2023overview}. Compared with traditional parameter estimation methods developed for particle filters, differentiable particle filters often require less knowledge about the considered state-space model~\cite{jonschkowski18,karkus2018particle,chen2021differentiable,corenflos2021differentiable,gama2023unsupervised}.

Most existing differentiable particle filtering frameworks are trained offline in a supervised approach. In supervised offline training schemes, ground-truth latent states are required for model training; subsequently, online inference is performed on new data without further updates of the model. This leads to several limitations. For example, in many real-world Bayesian filtering problems, data often arrive in a sequential order and ground-truth latent states can be expensive to obtain or even inaccessible. Moreover, offline training schemes can produce poor filtering results in the testing stage if the distribution of the offline training data differs from the distribution of the online data in testing, a.k.a. distribution shift.

Traditional online parameter estimation methods for particle filters are often designed as a nested structure for solving two layers of Bayesian filtering problems to simultaneously track the posterior of model parameters and latent variables. In~\cite{andrieu2010particle}, it was proposed to estimate the parameters of particle filters using Markov chain Monte Carlo (MCMC) sampling. The $\text{SMC}^2$ algorithm~\cite{chopin2013smc2} and the nested particle filter~\cite{crisan2018nested} leverage a nested filtering framework where two Bayesian filters are running hierarchically to approximate the joint posterior of unknown parameters and latent states, while the nested particle filter is designed as a purely recursive method thus more suitable for online parameter estimation problems. The scope of their applications is limited, since they assume that the structure of the considered state-space model is known.

To the best of our knowledge, no existing differentiable particle filters address online training at testing time. Several differentiable particle filters require ground truth state information in model training, rendering them unsuitable for online training. For example, in~\cite{jonschkowski18,karkus2018particle,chen2022conditional,corenflos2021differentiable}, differentiable particle filters are optimised by minimising the distance between the estimated latent states and the ground-truth latent states, e.g. the root mean square error (RMSE) or the negative log-likelihood loss. It was proposed in~\cite{gama2023unsupervised} to learn the sampling distributions of particle filters in an unsupervised manner. All of these methods were proposed for training differentiable particle filters on fixed, offline datasets.

In this paper, we introduce online learning differentiable particle filters (OL-DPFs). In OL-DPFs, parameters of differentiable particle filters are optimised by maximising a training objective modified from the evidence lower bound (ELBO) of the marginal log-likelihood in an online, unsupervised manner. Upon the arrival of new observations, the proposed OL-DPF is able to simultaneously update its estimates of the latent posterior distribution and model parameters.
Compared with previous online parameter estimation approaches developed for particle filters, we do not assume prior knowledge of the structure of the considered state-space model. Instead, we construct the components of particle filters using expressive neural networks such as normalising flows~\cite{chen2021differentiable,chen2022conditional}.

The rest of the paper is organised as follows. Section~\ref{sec:problem} gives a detailed introduction to the problem we consider in this paper. Section~\ref{sec:preliminaries} introduces background knowledge on normalising flow-based differentiable particle filters that we employ in the experiment section. The proposed online learning framework for differentiable particle filters is presented in Section~\ref{sec:methodology}. We evaluate the performance of the proposed method in numerical experiments and report the experimental results in Section~\ref{sec:experiment}. We conclude the paper in Section~\ref{sec:conclusion}.


\section{Problem Statement} \label{sec:problem}
In this work, we consider a state-space model that has the following form:
\begin{align}
	\label{eq:state_spaces_model}
	&x_0 \sim \pi\left(x_0\right)\,, 
	\\ \label{dyn} 
	&x_t \sim p\left(x_t | x_{t-1};\theta\right)\,,\,t\geq 1\,,
	\\ \label{mea} 
	&y_t \sim p\left(y_t | x_t;\theta\right)\,,\,t\geq 1\,,
\end{align}
where the latent state variable $({x}_t)_{t\geq0}$ is defined on $\mathcal{X}\subseteq\mathbb{R}^{d_\mathcal{X}}$, and the observed measurement variable $({y}_t)_{t\geq1}$ is defined on $\mathcal{Y}\subseteq\mathbb{R}^{d_\mathcal{Y}}$. $\pi(x_0)$ is the initial distribution of the latent state at time step $0$, $p(x_t|x_{t-1};\theta)$ is the dynamic model that describes the evolution of the latent state,
$p(y_t|x_t;\theta)$ is the measurement model that
specifies the relation between observations and latent states, $\theta$ refers to the parameters of the state-space model. In addition, we use $q(x_{t}| y_t, x_{t-1};\phi)$ to denote proposal distributions used to generate new particles in particle filtering approaches.

This paper addresses the problem of jointly estimating $\theta$, $\phi$, and latent posteriors $p(x_{0:t}|y_{1:t};\theta)$ or $p(x_{t}|y_{1:t};\theta)$ given the sequence of observations $y_{1:t}$ through an online approach, where $x_{0:t}:= \{x_0, x_{1}, \cdots, x_{t}\}$ and $y_{1:t}:= \{y_1, y_{2}, \cdots, y_{t}\}$. Following the practice of iterative filtering approaches proposed in~\cite{ionides2006inference,ionides2015inference}, we approximate the ground-truth state space model in the testing stage with a time-varying model, in which the constant model parameter set $\theta$ and proposal parameter set $\phi$ are replaced by a time-varying process $\theta_t$ and $\phi_t$, respectively. Our problem is hence converted to simultaneously track the joint posterior distribution
$p(x_{0:t}|y_{1:t};\theta_t)$ or the marginal posterior distribution $p(x_t|y_{1:t};\theta_t)$ and update the parameter sets $\theta_t$ and $\phi_t$ as $y_t$ arrives.



\section{Preliminaries}
\label{sec:preliminaries}
\subsection{Particle filtering}
\label{subsec:pf}
In particle filters, the latent posterior distribution $p(x_{0:t}| y_{1:t};\theta)$ is approximated with an empirical distribution:
	\begin{gather}
		\label{eq:pf_approx}
		p(x_{0:t}| y_{1:t};\theta)\approx\sum_{i=1}^{N_p}\textbf{w}_t^i\, \delta_{x_{0:t}^i}(x_{0:t})\,,
        \textbf{w}_t^i=\frac{w_t^i}{\sum_{j=1}^{N_p}w_t^j}\,,
	\end{gather}
	where $N_p$ is the number of particles, $\delta_{x_{0:t}^i}(\cdot)$ denotes the Dirac delta function located in $x_{0:t}^i$, $\textbf{w}_t^i\geq 0$ with $\sum_{i=1}^{N_p}\textbf{w}_t^i=1$ is the normalised importance weight of the $i$-th particle at the $t$-th time step, and $w_t^i$ refers to unnormalised particle weights.

The particles $x_{0:t}^i$, ${i\in \{1,2,\cdots,N_p\}}$ are sampled from the initial distribution $\pi(x_{0})$ when $t=0$ and proposal distributions $q(x_{t}| y_t, x_{t-1};\phi)$ for $t\geq1$. When a predefined condition is satisfied, e.g. the effective sample size (ESS) is lower than a threshold, particle resampling is performed to reduce particles with small weights~\cite{douc2005comparison}.

Denote by $A_t^i$ ancestor indices in resampling, unnormalised importance weights $w_{t+1}^i$ are updated as follows:
	\begin{equation}
		\label{eq:weight_update}
		w_{t+1}^i= \tilde{w}_{t}^i \frac{p(y_{t+1}| x_{t+1}^i;\theta)p(x_{t+1}^i| \tilde{x}_{t}^i;\theta)}{q(x_{t+1}^i| y_{t+1}, \tilde{x}_{t}^i;\phi)}\,,
	\end{equation}
	where $w_{0}^i=1$, $\tilde{w}_{t}^i$ refers to unnormalised weights after resampling ($\tilde{w}_{t}^i=1$ if resampled at $t$, otherwise $\tilde{w}_{t}^i={w}_{t}^i$), and $\tilde{x}_{t}^i$ denote particle values after resampling ($\tilde{x}_{t}^i={x}_{t}^{A_t^i}$ if resampled at $t$, otherwise $\tilde{x}_{t}^i={x}_{t}^i$).
    
\subsection{Normalising flow-based particle filters}
\label{subsec:nf-dpf}
Since we assume that the functional form of state-space models we consider is unknown, we construct the proposed OL-DPF with a flexible differentiable particle filtering framework called normalising flows-based differentiable particle filters (NF-DPFs)~\cite{chen2021differentiable, chen2022conditional}. The components of the NF-DPFs are constructed as follows.
\subsubsection{Dynamic model with normalising flows}
Suppose that we have a base distribution $g\left(\cdot | x_{t-1} ; \theta\right)$ which follows a simple distribution such as a Gaussian distribution. A sample ${x}_t^i$ from the dynamic model of NF-DPFs~\cite{chen2021differentiable} can be obtained by applying a normalising flow $\mathcal{T}_\theta(\cdot): \mathcal{X} \rightarrow \mathcal{X}$ to a particle $\dot{x}_t^i$ drawn from the base distribution $g\left(\cdot | x_{t-1} ; \theta\right)$:
\begin{gather}
\dot{x}_t^i \sim g\left(\dot{x}_t | x_{t-1} ; \theta\right), \\
{x}_t^i=\mathcal{T}_\theta\left(\dot{x}_t^i\right) \sim p\left({x}_t | x_{t-1} ; \theta\right) .
\end{gather}
The probability density of a given ${x}_t$ in the dynamic model can be obtained by applying the change of variable formula:
\begin{gather}
p\left({x}_t | x_{t-1} ; \theta\right)=g\left(\dot{x}_t | x_{t-1} ; \theta\right)\left|\operatorname{det} J_{\mathcal{T}_\theta}\left(\dot{x}_t\right)\right|^{-1}, \\
\dot{x}_t=\mathcal{T}_\theta^{-1}\left({x}_t\right) \sim g\left(\dot{x}_t | x_{t-1} ; \theta\right),
\end{gather}
where $\operatorname{det}J_{\mathcal{T}_\theta}(\dot{x}_t)$ is the Jacobian determinant of $\mathcal{T}_\theta(\cdot)$ evaluated at $\dot{x}_t=\mathcal{T}_\theta^{-1}({x}_t)$.
\subsubsection{Proposal distribution with conditional normalising flows}
Samples from the proposal distribution of NF-DPFs ~\cite{chen2021differentiable} can be obtained by applying a conditional normalising flow $\mathcal{F}_\phi(\cdot): \mathcal{X} \times \mathcal{Y} \rightarrow \mathcal{X}$ to samples drawn from a base proposal distribution $h\left(\cdot | x_{t-1}, y_t ; \phi\right)$, $ t \geq 1$:
\begin{gather}
\hat{x}_t^i \sim h\left(\hat{x}_t | x_{t-1}, y_t ; \phi\right), \\
x_t^i=\mathcal{F}_\phi\left(\hat{x}_t^i ; y_t\right) \sim q\left(x_t | x_{t-1}, y_t ; \phi\right)\,,
\end{gather}
where the conditional normalising flow $\mathcal{F}_\phi(\cdot)$ is an invertible function of particles $\hat{x}_t^i$ given the observation $y_t$.

By applying the change of variable formula, the proposal density of ${x}_t$ can be computed as:
\begin{gather}
q\left(x_t | x_{t-1}, y_t ; \phi\right)=h\left(\hat{x}_t | x_{t-1}, y_t ; \phi\right)\left|\operatorname{det} J_{\mathcal{F}_\phi}\left(\hat{x}_t ; y_t\right)\right|^{-1} \,.\\
\hat{x}_t=\mathcal{F}_\phi^{-1}\left(x_t ; y_t\right) \sim h\left(\hat{x}_t | x_{t-1}, y_t ; \phi\right),
\end{gather}
where $\operatorname{det} J_{\mathcal{F}_\phi}\left(\hat{x}_t ; y_t\right)$ refers to the determinant of the Jacobian matrix $J_{\mathcal{F}_\phi}\left(\hat{x}_t ; y_t\right)=\frac{\partial \mathcal{F}_\phi\left(\hat{x}_t ; y_t\right)}{\partial \hat{x}_t}$ evaluated at $\hat{x}_t^i$.
\subsubsection{Measurement model with conditional normalising flows}
In NF-DPFs, the relation between the observation
and the state is constructed with another conditional normalising flow $\mathcal{G}_\theta(\cdot): \mathbb{R}^{d_{y}} \times \mathcal{X}: \rightarrow \mathcal{Y}$~\cite{chen2022conditional}:
\begin{gather}
y_t=\mathcal{G}_\theta\left(z_t ; x_t\right) \label{measurement cnf}
\end{gather}
where $z_t = \mathcal{G}_\theta^{-1}\left(y_t ; x_t\right)$ is the base variable which follows a user-specified independent marginal distribution $p_Z\left(z_t\right)$ defined on $\mathbb{R}^{d_{y}}$ such as an isotropic Gaussian. By modelling the generative process of $y_t$ in such a way, the likelihood of the observation $y_t$ given $x_t$ can be computed by:
\begin{gather}
p\left(y_t | x_t ; \theta\right)=p_Z\left(z_t\right)\left|\operatorname{det} J_{\mathcal{G}_\theta}\left(z_t ; x_t\right)\right|^{-1}\,,
\end{gather}
where $z_t=\mathcal{G}_\theta^{-1}\left(y_t ; x_t\right)$, and $\operatorname{det} J_{\mathcal{G}_\theta}\left(z_t ; x_t\right)$ denotes the determinant of the Jacobian matrix $J_{\mathcal{G}_\theta}\left(z_t ; x_t\right)=\frac{\partial \mathcal{G}_\theta\left(z_t ; x_t\right)}{\partial z_t}$ evaluated at $z_t^i$.

A pseudocode that describes how to propose new particles and update weights in NF-DPFs is provided in Algorithm~\ref{alg:online_nf}.

\begin{algorithm}[htbp]
	\begin{algorithmic}[1]
		\caption{Proposal and weight update step of a normalising flow-based differentiable particle filter.\\(Can be used to replace line 12-13 of Algorithm \ref{alg:online_general}.)}
		\label{alg:online_nf}
            \STATE \textbf{Notations:}\\
            \begin{footnotesize}	
				\hspace{-2em}\begin{tabular}[t]{l @{\hspace{.02em}} l}%
					$g(\cdot;\theta)$ & Base dynamic model\\
                        $h(\cdot;\phi)$ & Base proposal distribution\\
					$p_Z(\cdot)$ & Standard Gaussian PDF\\
                    
				\end{tabular}\hspace{-0.5em}%
				\begin{tabular}[t]{l @{\hspace{.1em}} l}
					$\mathcal{T}_\theta(\cdot)$ & Dynamic normalising flow\\
					$\mathcal{F}_\phi(\cdot)$ & Proposal normalising flow\\
					$\mathcal{G}_\theta(\cdot)$ & Measurement normalising flow
				\end{tabular}
			\end{footnotesize}
            \vspace{0.05em}
            \STATE \textbf{Input:}
					$\tilde{x}_{t-1}^i$: particles at $t$-1 after resampling,
					$\tilde{w}_{t-1}^i$: weights at $t$-1 after resampling,
					$y_{t}$: observations at $t$;
            \vspace{0.1em}
            \STATE Sample $\hat{x}_t^i \stackrel{\text { i.i.d. }}{\sim} h\left(\hat{x}_t | \tilde{x}^i_{t-1}, y_t ; \phi\right)$;
		\STATE Generate proposed particles \\$x_t^i=\mathcal{F}_{\phi}\left(\hat{x}_t^i ; y_t\right) \sim q\left(x_t|\tilde{x}^i_{t-1}, y_t ; \phi\right)$;
            \STATE Compute the base variable $z_t^i=\mathcal{G}_{\theta}^{-1}\left(y_t ; x_t^i\right)$;
		\STATE Compute and update importance weight \\$w_t^i=\tilde{w}_{t-1}^i \frac{p_Z\left(z_t^i\right)\left|\operatorname{det} J_{\mathcal{F}_{\phi}}\left(\hat{x}_t^i ; y_t\right)\right| g\left(\dot{x}_t^i | \tilde{x}^i_{t-1}; \theta\right)}{\left|\operatorname{det} J_{\mathcal{G}_{\theta}}\left(z_t^i ; x_t^i\right)\right| h\left(\hat{x}_t^i | \tilde{x}^i_{t-1}, y_t ; \phi\right)\left|\operatorname{det} J_{\mathcal{T}_{\theta}}\left(\dot{x}_t^i\right)\right|}$;
  \vspace{0.5em}
  \STATE \textbf{Output:} $x_t^i$: proposed particles,  $w_t^i$: updated weights.
	\end{algorithmic}
\end{algorithm}

\section{Online learning framework with unsupervised training objective} \label{sec:methodology}
In this section, we present details of the proposed online learning differentiable particle filters (OL-DPFs), including the training objective we adopt to train OL-DPFs.

\subsection{Unsupervised online learning objective} \label{sec:objective}
Given a set of observations $y_{1: T}$, one can optimise the model parameters $\theta$ and the proposal parameters $\phi$ in an unsupervised way by maximising an approximation to the filtering evidence lower bound (ELBO) of the marginal log-likelihood~\cite{le2018auto,maddison2017filtering,naesseth2018variational}. This filtering ELBO is derived based on the unbiased estimator $\hat{p}(y_{1: T};\theta)$ of the marginal likelihood ${p}(y_{1: T};\theta)$ obtained from particle filters~\cite{le2018auto,maddison2017filtering,naesseth2018variational}:
\begin{align}
	\log\;{p}(y_{1: T};{\theta}) & = \log\;\mathbb{E}\Big[\hat{p}(y_{1: T};\theta)\Big] \label{eq:eq1} \\
	& \geq\mathbb{E}\Big[\log\;\hat{p}(y_{1: T};\theta)\Big] \label{eq:eq2}\,,
\end{align}
where Jensen's inequality is applied from Eq.~\eqref{eq:eq1} to Eq.~\eqref{eq:eq2}. An unbiased estimator of the ELBO $\mathbb{E}\Big[\log\;\hat{p}(y_{1: T};\theta)\Big]$ can be obtained by computing:
\begin{gather}
\label{eq:smc_estimator}
	\sum_{t=1}^{T}\log\left[\sum_{i=1}^{N_p}\frac{ {w}_t^i}{\sum_{j=1}^{N_p} \tilde{w}_{t-1}^j} \right],
\end{gather}
where  $\tilde{w}_{t}^i$ refers to unnormalised weights after resampling ($\tilde{w}_{t}^i=1$ if resampled at $t$, otherwise $\tilde{w}_{t}^i={w}_{t}^i$)~\cite{chopin2020particle}.

However, in online learning settings, directly maximising Eq.~\eqref{eq:smc_estimator} for the optimisation of $\theta$ and $\phi$ is often impractical. One reason is that the computation of Eq.~\eqref{eq:smc_estimator} does not scale in online learning settings as $T$ grows. 
As an alternative solution, we propose to decompose the whole trajectory into sliding windows of length $L$ and update $\theta$ and $\phi$ every $L$ time steps. Denote by $S\in\{0,1,\cdots\}$ the indices of sliding windows, we use $\theta_S$ and $\phi_S$ to denote the model parameters and proposal parameters in the $S$-th sliding window, which are kept constant between time steps $SL+1$ to $(S+1)L$. The initial values $\theta_0$ and $\phi_0$ are set to the pre-trained model parameters $\theta^*$ and proposal parameters $\phi^*$ obtained from the offline training stage. The $S$-th sliding window contains $L$ observations $y_{SL+1:(S+1)L}$. 

For the $(S+1)$-th sliding window, the goal is to update $\theta_S$ and $\phi_S$ to $\theta_{S+1}$ and $\phi_{S+1}$ by maximising an ELBO on the conditional log-likelihood of observations $\log p(y_{SL+1:(S+1)L}|y_{1:SL};\theta_S)$:
\begin{align}
	&\log\;{p}(y_{SL+1:(S+1)L}|y_{1:SL};\theta_S) \\=& \log\;\mathbb{E}\Big[\hat{p}(y_{SL+1:(S+1)L}|y_{1:SL};\theta_S)\Big] \\
	 \geq&\mathbb{E}\Big[\log\;\hat{p}(y_{SL+1:(S+1)L}|y_{1:SL};\theta_S)\Big]\label{eq:elbo_conditional}\\
    =&\mathbb{E}\left[\sum_{t=SL+1}^{(S+1)L}\log\left[\sum_{i=1}^{N_p}\frac{ \dot{w}_t^i}{\sum_{j=1}^{N_p} \dot{\tilde{w}}_{t-1}^j} \right]\right]\,,
\end{align}
where $\dot{w}_t^i$ and $\dot{\tilde{w}}_t^i$ are unnormalised weights before and after resampling at time step $t$. Both weights are computed with fixed parameters $\theta_S$ and $\phi_S$ up to the current time step $(S+1)L$, such that $\hat{p}(y_{SL+1:(S+1)L}|y_{1:SL};\theta_S)$ is an unbiased estimator of ${p}(y_{SL+1:(S+1)L}|y_{1:SL};\theta_S)$. However, the computation of $\hat{p}(y_{SL+1:(S+1)L}|y_{1:SL};\theta_S)$ is computationally expensive for large $S$ because it requires rerunning the filtering algorithm from time step 0. Therefore, we approximate $\hat{p}(y_{SL+1:(S+1)L}|y_{1:SL};\theta_S)$ by using particle weights produced with time-varying parameters $\{\theta_k\}_{k=0}^{S}$ and $\{\phi_k\}_{k=0}^{S}$:
\begin{gather}
\label{eq:approx_conditional}
    \prod_{t=SL+1}^{(S+1)L}\left[\sum_{i=1}^{N_p}\frac{ {w}_t^i}{\sum_{j=1}^{N_p} \tilde{w}_{t-1}^j} \right]\,.
\end{gather}
The full details of the proposed OL-DPF approach, including the computation of ${w}_t^i$ and $\tilde{w}_{t-1}^j$, are presented in Algorithm~\ref{alg:online_general}. Note that Eq.~\eqref{eq:approx_conditional} represents a coarse estimation of $\hat{p}(y_{SL+1:(S+1)L}|y_{1:SL};\theta_S)$ due to the presence of time-varying parameters.

By incorporating the approximation given by Eq.~\eqref{eq:approx_conditional} into the ELBO defined by Eq.~\eqref{eq:elbo_conditional}, the overall loss function that OL-DPFs employ to update $\theta_S$ and $\phi_S$ is as follows:
\begin{gather}
\label{eq:loss}
	\mathcal{L}(\theta_S, \phi_S)=-\sum_{t=SL+1}^{(S+1)L}\log\left[\sum_{i=1}^{N_p}\frac{ {w}_t^i}{\sum_{j=1}^{N_p} \tilde{w}_{t-1}^j} \right].
\end{gather}

    \begin{algorithm}[htbp]
	\begin{algorithmic}[1]
		\caption{OL-DPF (General).}
		\label{alg:online_general}
		
		\STATE \textbf{Notations:}\\ 
            \begin{footnotesize}	
				\hspace{-2em}\begin{tabular}[t]{l @{\hspace{0.5em}} l}%
					$\text{ESS}_\text{thres}$ & resampling threshold \\
                        $\alpha$ & learning rate \\
					$\pi\left(x_0\right)$ & initial distribution\\
                    
				\end{tabular}\hspace{-0.5em}%
				\begin{tabular}[t]{l @{\hspace{.1em}} l}
					$N_p$ & number of particles\\
					$L$ & length of sliding windows\\
                        $\textbf{Mult}(\cdot)$& multinomial distribution
				\end{tabular}
			\end{footnotesize}
            \vspace{0.5em}
            
		\STATE \textbf{Initialise} $\theta$ and $\phi$ randomly;
		\STATE \textbf{Pre-train} $\theta$ and $\phi$ with offline training data using a supervised loss until $\theta$ and $\phi$ converge to ${\theta}^*$ and ${\phi}^*$;
		\STATE \textbf{Online learning:} 
            \STATE Sliding window index $S=0$;
            \STATE $\theta_0 \leftarrow {\theta}^*$, $\phi_0 \leftarrow {\phi}^*$;
		\STATE Draw particles $\left\{x_0^i\right\}_{i=1}^{N_p}$ from the $\pi\left(x_0\right)$; 
		\STATE Set importance weights $\left\{w_0^i\right\}_{i=1}^{N_p}=\frac{1}{N_p}$, $\left\{\tilde{w}_0^i\right\}_{i=1}^{N_p}=\frac{1}{N_p}$; 
		\STATE $t=1$;
		\WHILE{\textbf{online-learning} not completed}
		
		
		\FOR{$i=1$ to $N_p$}
		\STATE Draw particles from $q\left(x_t^i | \tilde{x}_{t-1}^i, y_t ; \phi_S\right)$;
		\STATE Update particle weight \\$w_t^i=\tilde{w}_{t-1}^i \frac{p\left(x_t^i | \tilde{x}_{t-1}^i ; \theta_S\right) p\left(y_t | x_t^i ; \theta_S\right)}{q\left(x_t^i | \tilde{x}_{t-1}^i, y_t ; \phi_S\right)}$;
		\ENDFOR
		\STATE Normalize weights $\left\{\textbf{w}_t^i=\frac{w_t^i}{\sum_{n=1}^{N_p} w_t^n}\right\}_{i=1}^{N_p}$;
		\STATE Estimate the hidden state $\hat{x}_t=\sum_{i=1}^{N_p} \textbf{w}_t^{i} x_t^{i}$;
		\IF{$t\bmod L==0$}
		\STATE Compute $\mathcal{L}(\theta_S, \phi_S)$ using Eq.~\eqref{eq:loss};
		\STATE Update $\theta$ and $\phi$ through gradient descent: \\$\theta_{S+1} \leftarrow \theta_S-\alpha \nabla_\theta \mathcal{L}(\theta_S, \phi_S)$, \\$\phi_{S+1} \leftarrow \phi_S-\alpha \nabla_\theta \mathcal{L}(\theta_S, \phi_S)$;
            \STATE $S=S+1$;
		\ENDIF
		\STATE Compute the effective sample size: $\mathrm{ESS}_t=\frac{1}{\sum_{i=1}^{N_p}\left(\textbf{w}_t^{i}\right)^2}$;
		\IF{$\text{ESS}_t<\text{ESS}_\text{thres}$}
		\STATE Sample $A_t^i\sim\textbf{Mult}(\textbf{w}_t^1,\cdots,\textbf{w}_t^{N_p})$ for $\forall i$ to obtain $\left\{\tilde{w}_{t}^i=1\right\}_{i=1}^{N_p}, \left\{\tilde{x}_t^{i} = x_t^{A_t^i}\right\}_{i=1}^{N_p}$;
            \ELSE 
            \STATE $\left\{\tilde{w}_{t}^i={w}_{t}^i\right\}_{i=1}^{N_p}$, $\left\{\tilde{x}_t^{i} = x_t^{i}\right\}_{i=1}^{N_p}$;
		\ENDIF
		\STATE $t=t+1$;
		\ENDWHILE
	\end{algorithmic}
\end{algorithm}

\section{Experiment setup and results} 
\label{sec:experiment}
In this section, we evaluate the performance of the proposed OL-DPFs in two simulated numerical experiments. We first consider in Section~\ref{multi_linear} a parameterised multivariate linear Gaussian state-space model with varying dimensionalities and ground-truth parameter values. The structure of the state-space model follows the setup adopted in~\cite{naesseth2018variational,corenflos2021differentiable}. We then validate the effectiveness of the proposed OL-DPFs in a non-linear position tracking task, where the dynamics of the tracked object are described by a Markovian switching dynamic model~\cite{bugallo2007performance}. 


In both experiments, we pre-train the OL-DPF by minimising a supervised lose, i.e. (root) mean square errors between the estimated states and the ground-truth latent states available in the offline data. In the online learning (testing) stage, the OL-DPF is optimised by minimising the loss function specified in Eq.~\eqref{eq:loss}.
Two baselines are considered in this section. The first baseline is the pre-trained DPF, which is only optimised in the pre-training stage and the model parameters remain fixed in the online learning stage. The second baseline, referred as the DPF in this section, is trained with supervised losses in both the pre-training and online learning stages as a gold standard benchmark. RMSEs produced by different approaches are used to compare the performance of different methods. In all experiments, the Adam optimiser~\cite{kingma2015adam} is used to perform gradient descent, and the number of particles and the learning rate are set to 100 and 0.005, respectively\footnote{Code is available at: https://github.com/JiaxiLi1/Online-Learning-Differentiable-Partiacle-Filters.}.


\subsection{Multivariate linear Gaussian state-space model} \label{multi_linear}
\subsubsection{Experiment setup}
We first consider a multivariate linear Gaussian state-space model formulated in~\cite{naesseth2018variational,corenflos2021differentiable}:
\begin{gather}
	\label{linear gaussian}
	x_0 \sim \mathcal{N}\left(\mathbf{0}_{d_{\mathcal{X}}}, \mathbf{I}_{d_{\mathcal{X}}}\right), \\
	x_t | x_{t-1} \sim \mathcal{N}\left(\tilde{\boldsymbol{\theta}}_1 x_{t-1}, \mathbf{I}_{d_{\mathcal{X}}}\right), \\
	y_t | x_t \sim \mathcal{N}\left(\tilde{\boldsymbol{\theta}}_2 x_t, 0.1\mathbf{I}_{d_{\mathcal{X}}}\right)\,,
\end{gather}
where $\mathbf{0}_{d_{\mathcal{X}}}$ is a $d_{\mathcal{X}} \times d_{\mathcal{X}}$ null matrix, $\mathbf{I}_{d_{\mathcal{X}}}$ is a $d_{\mathcal{X}} \times d_{\mathcal{X}}$ identity matrix. $\tilde{\boldsymbol{\theta}}:=\left(\tilde{\boldsymbol{\theta}}_1\in\mathbb{R}^{d_{\mathcal{X}}\times d_{\mathcal{X}}}, \tilde{\boldsymbol{\theta}}_2\in \mathbb{R}^{d_{\mathcal{Y}}\times d_{\mathcal{X}}}\right)$ is the model parameters. Our target is to track the hidden state $x_t$ given observation data $y_t$. In this experiment, we set $d_{\mathcal{X}}=d_{\mathcal{Y}}$, i.e. observations and latent states are of the same dimensionality. We compare the performance of different approaches for $d_{\mathcal{X}} \in\{2,5,10\}$.

We set different values for $\tilde{\boldsymbol{\theta}}$, denoted by $\tilde{\boldsymbol{\theta}}_{\text{pre-train}}:=\left(\tilde{\boldsymbol{\theta}}_{1_\text{pre-train}}, \tilde{\boldsymbol{\theta}}_{2_\text{pre-train}}\right)$ and $\tilde{\boldsymbol{\theta}}_{\text{online}}:=\left(\tilde{\boldsymbol{\theta}}_{1_\text{online}}, \tilde{\boldsymbol{\theta}}_{2_\text{online}}\right)$ respectively for the pre-training stage and the online learning stage to generate distribution shift data. Specifically, for $\tilde{\boldsymbol{\theta}}_{\text{pre-train}}$, we set the element of $\tilde{\boldsymbol{\theta}}_{1_{\text{pre-train}}}$ at the intersection of its $i$-th row and $j$-th column as $\tilde{\boldsymbol{\theta}}_{1_{\text{pre-train}}}(i, j)=\left(0.42^{|i-j|+1}\right)$, $1 \leq i, j \leq d_{\mathcal{X}}$, and $\tilde{\boldsymbol{\theta}}_{2_{\text{pre-train}}}$ is a diagonal matrix with 0.5 on its diagonal line. For the online learning stage, we set $\tilde{\boldsymbol{\theta}}_{1_{\text{online}}}(i, j)=\left(0.2^{|i-j|+1}\right)$, $1 \leq i, j \leq d_{\mathcal{X}}$, and $\tilde{\boldsymbol{\theta}}_{2_{\text{online}}}$ is a diagonal matrix with 10.0 on the diagonal line.

The pre-training data contains 500 trajectories. There are 50 time steps in each trajectory. The test-time data is a trajectory with 5,000 time steps. Particles at time step $0$ are initialised uniformly over the hypercube whose boundaries are defined by the minimum and the maximum values of ground-truth latent states in the offline training data at each dimension. The length of sliding windows $L$ are set to be 10 for the online learning stage.  

\subsubsection{Experiment results}
\begin{figure}[!ht]
	\centering



	\begin{subfigure}[t]{\linewidth}
        \caption{Dimension=2}
        \includegraphics[width=\textwidth]{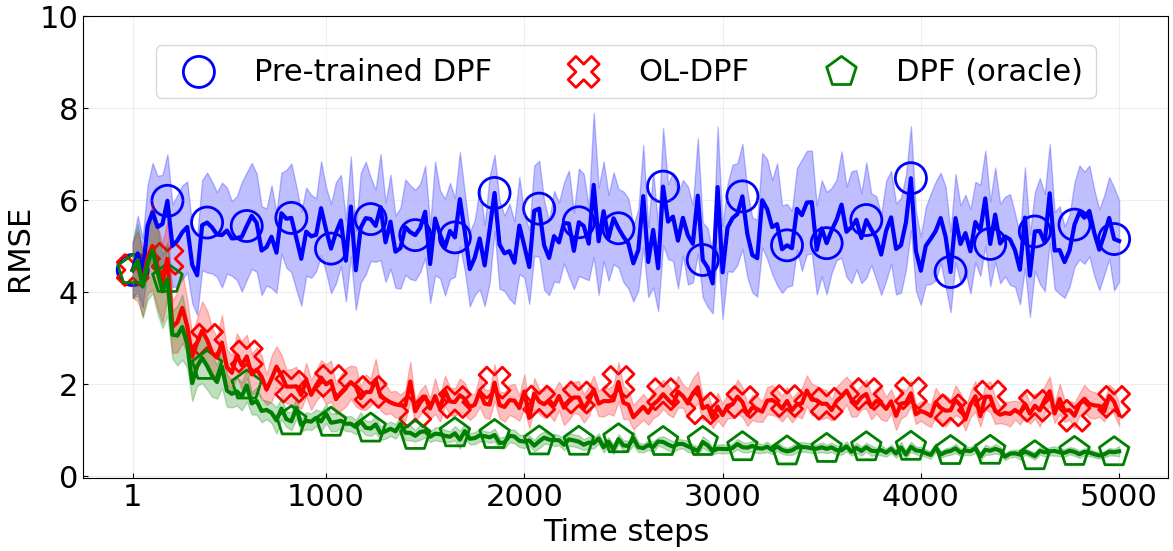}
	\end{subfigure}\par\vspace{0\baselineskip}

	\begin{subfigure}[t]{\linewidth}
 \caption{Dimension=5}
        \includegraphics[width=\textwidth]{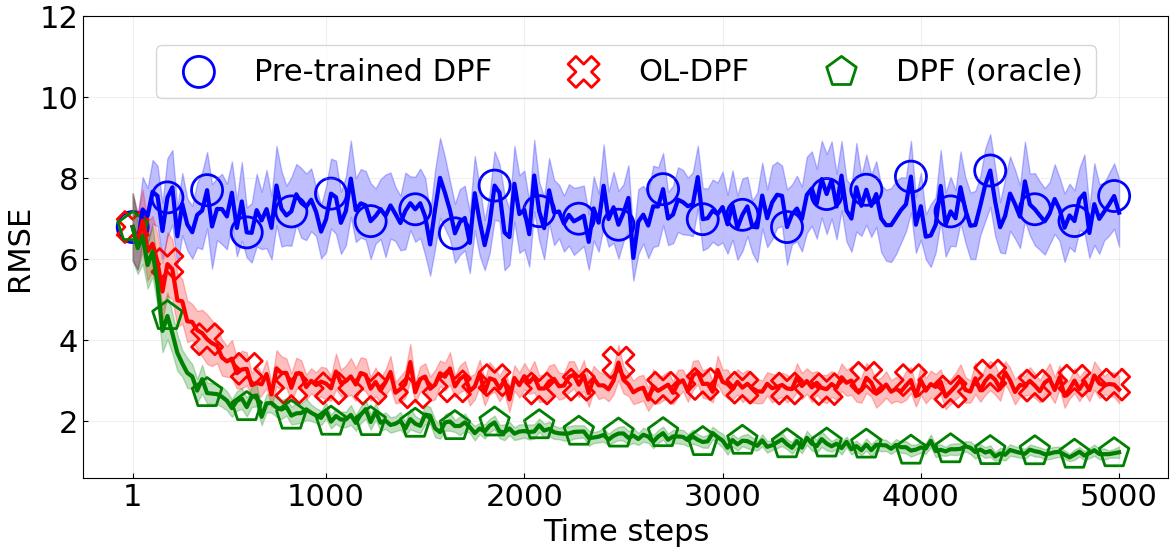}
	\end{subfigure}\par\vspace{0\baselineskip}

	\begin{subfigure}[t]{\linewidth}
 \caption{Dimension=10}
        \includegraphics[width=\textwidth]{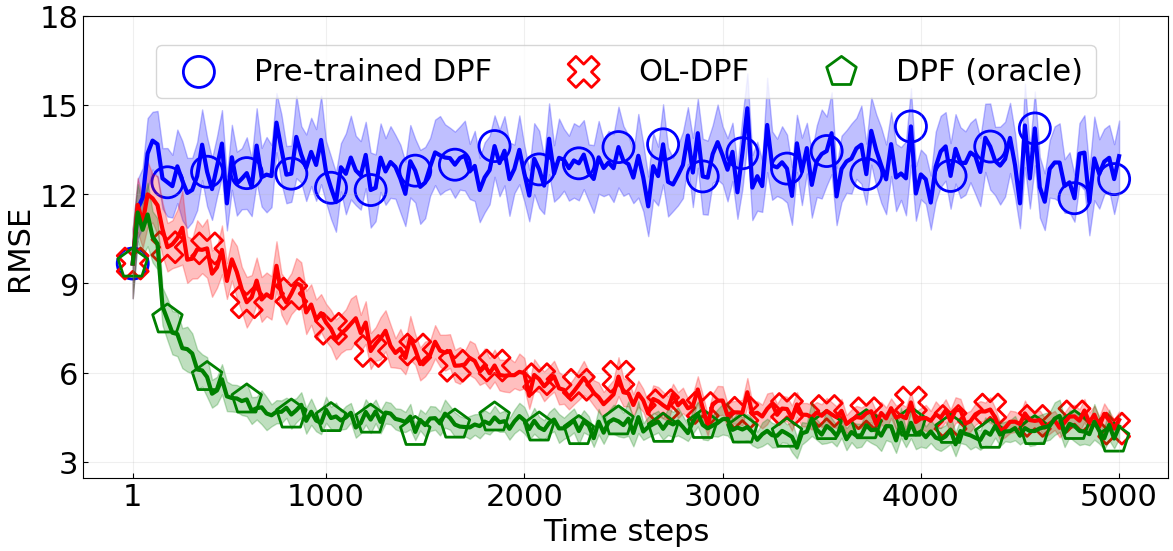}
	\end{subfigure}\par\vspace{0\baselineskip}

	\caption{The mean and confidence intervals of test RMSEs produced by different methods in the online learning stage in $d_\mathcal{X}$-dimensional multivariate linear Gaussian state-space models, with $d_\mathcal{X}\in\{2,5,10\}$. The shaded area represents the $95\%$ confidence interval computed among 50 random simulations. 
	}
	\label{fig1}
\end{figure}

The test RMSEs in the online learning stage produced by the evaluated methods are presented in Fig. \ref{fig1}. The RMSE of the OL-DPF converge to a stable value within 1,000 time steps in 2-dimensional and 5-dimensional cases. It took around 3,000 time steps to converge in the 10-dimensional experiment, likely caused by the increasing number of learnable parameters as $d_\mathcal{X}$ grows. Table~\ref{tab1} provides the mean and standard deviation of RMSEs of different methods computed over 50 random runs in the online learning phase. The DPF achieves the lowest RMSE among all the evaluated methods as expected, since DPFs are trained with ground-truth latent states in both the offline stage and the online stage. Compared with the pre-trained DPF, the OL-DPF not only yields lower estimation errors but also presents lower standard deviations in all three experiment setups.

\begin{table}[htbtp]
	\centering
	\caption{RMSEs in the online learning stage of different methods in the multivariate linear Gaussian experiment. The reported mean and standard deviation are computed with 50 random simulations.}
	\label{tab1}  
	\begin{tabular}{|c|c|c|c|}
		
		\hline
		Method     & RMSE$_{d_\mathcal{X}=2}$   &RMSE$_{d_\mathcal{X}=5}$ & RMSE$_{d_\mathcal{X}=10}$  \\ \hline

  		Pre-trained DPF & $5.30 \pm 3.51$  & $7.21 \pm 2.86$ & $12.94 \pm 3.94$  \\ \hline

		 OL-DPF    & $1.83 \pm 1.47$  & $3.14 \pm 1.49$ & $6.12 \pm 2.82$  \\ \hline
		
		DPF (oracle)    & $1.00 \pm 1.23$  &  $1.88 \pm 1.21$&  $4.60 \pm 2.10$ \\ \hline
		
	\end{tabular}

\end{table}

\subsection{Non-linear object tracking} \label{non-linear}
\subsubsection{Experiment setup}
In this experiment, we assess the efficacy of the proposed online learning framework in a non-linear object tracking task. The dynamic of the tracked object is described as follows:
\begin{gather}
	\label{position_tracking}
	x_t=\left(\begin{array}{ll}
		\tilde{x}_t & \omega_t
	\end{array}\right)^{\top}\,, 
\end{gather}
where $\tilde{x}_t=\left(x_{1, t}, x_{2, t} \dot{x}_{1, t} \dot{x}_{2, t}\right)^{\top} \in \mathbb{R}^4$ encompasses the Cartesian coordinates for target position ($x_{1, t}, x_{2, t}$) and velocity ($\dot{x}_{1, t} \dot{x}_{2, t}$), and evolves according to  $\tilde{x}_t=\mathbf{A}\left(\omega_{t-1}\right) \tilde{x}_{t-1}+\mathbf{B} \mathbf{u}_t$, where  $\mathbf{u}_t \sim \mathcal{N}\left(\mathbf{0}, 10^{-2} \mathbf{I}_2\right)$. The state transition matrices are as follows:
\begin{gather}
	\mathbf{A}\left(\omega_t\right)=\left(\begin{array}{cccc}
		1 & 0 & \frac{\sin \left(\omega_t T_s\right)}{\omega_t} & -\frac{1-\cos \left(\omega_t T_s\right)}{\omega_t} \\
		0 & 1 & -\frac{1-\cos \left(\omega_t T_s\right)}{\omega_t} & \frac{\sin \left(\omega_t T_s\right)}{\omega_t} \\
		0 & 0 & \cos \left(\omega_t T_s\right) & -\sin \left(\omega_t T_s\right) \\
		0 & 0 & \sin \left(\omega_t T_s\right) & \cos \left(\omega_t T_s\right)
	\end{array}\right)\,,
\end{gather}
\begin{gather}
\mathbf{B}=\left(\begin{array}{cc}
		\frac{T_s^2}{2} & 0 \\
		0 & \frac{T_s^2}{2} \\
		T_s & 0 \\
		0 & T_s
	\end{array}\right)\,,
\end{gather}
where $T_s=5$ denotes the sampling period (in seconds) and $\omega_t $ is defined by:
\begin{equation}
	\omega_t=\frac{a}{\sqrt{\dot{x}_{1, t-1}^2+\dot{x}_{2, t-1}^2}}+u_{\omega, t}\,,
\end{equation}
with $a$ being the maneuvering acceleration at time $t$ and $u_{\omega, t} \sim\mathcal{N}\left(0, 10^{-4}\right)$ the process noise. 

Observations $y_t\in\mathbb{R}^2$ in this experiment are generated through:
\begin{gather}
    y_t=h\left(x_t\right)+\mathbf{v}_t\,,
\end{gather}
where $\mathbf{v}_t \sim 0.7 \mathcal{N}\left(\mathbf{0}, 4 \cdot \mathbf{I}_2\right)+0.3 \mathcal{N}\left(\mathbf{0}, 25 \cdot \mathbf{I}_2\right)$ is the measurement noise, and the function $h(x_t):=(h_1\left(x_t\right), h_2\left(x_t\right))$ is defined through:
\begin{gather}
\label{eq:tracking_measurement}
	h_1\left(x_t\right)=10 \log _{10}\left(\frac{P_0}{\left\|\mathbf{r}-\mathbf{p}_t\right\|^\beta}\right),\\ \quad h_2\left(x_t\right)=\angle\left(\mathbf{p}_t-\mathbf{r}\right)\,,
\end{gather}
where $\mathbf{p}_t=\left(\begin{array}{ll}x_{1, t} & x_{2, t}\end{array}\right)^{\top} \in \mathbb{R}^2$ denotes the target position and $\|\mathbf{z}\|=\sqrt{\mathbf{z}^{\top} \mathbf{z}}$ refers to the norm of a vector $\mathbf{z}$. 
Following the setup in~\cite{bugallo2007performance}, in Eq.~\eqref{eq:tracking_measurement}, we set $P_0=1$, $\beta=2.0$, and fix the reference point at $\mathbf{r}=(2,2)$. 

When generating pre-training and online learning data, we initialise the position and velocity of the object with $(0, 0)$ and $(\frac{55}{\sqrt{2}},\frac{55}{\sqrt{2}})$ for both the pre-training and online learning data, respectively. 
We use different values of $a$ for generating the pre-training data and the online learning data to simulate distribution shift data. Specifically, we use $a=5.0$ when generating data used for pre-training, and set $a$ to be $-5.0$ when generating data for the testing stage. The pre-training data consist of 500 trajectories, each with 50 time steps. For each simulation, data for online learning include a single trajectory with 5,000 time steps. Particles at the first time step are initialised uniformly over the hypercube whose boundaries are defined by the minimum and the maximum values of ground-truth latent states in the offline training data at each dimension. The length of sliding windows $L$ is set to 10 in this experiment.

\subsubsection{Experiment results}
\begin{figure}[!ht]
	\centering

	\begin{subfigure}[t]{\linewidth}
        \includegraphics[width=\textwidth]{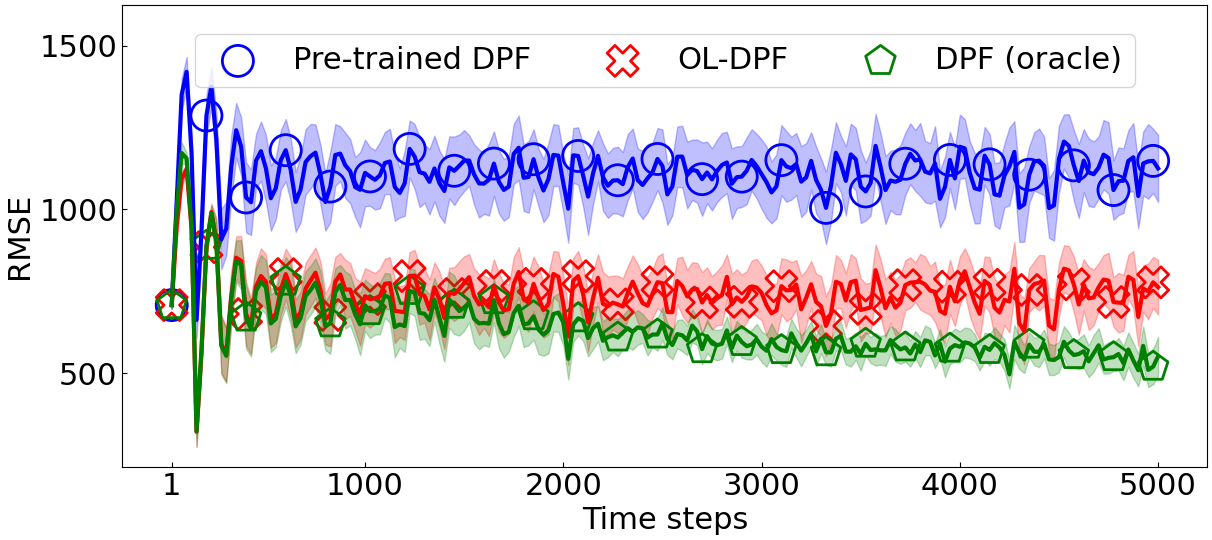}
	\end{subfigure}\par\vspace{0\baselineskip}

	\caption{The mean and the confidence interval of test RMSEs of different methods in the online learning stage. We report the mean and $95\%$ confidence interval of RMSEs among 50 random simulations. 
	}
	\label{fig2}
\end{figure}

Fig.~\ref{fig2} shows the mean and the standard deviation of test RMSEs produced by different method computed with 50 random seeds. RMSEs of OL-DPFs converge in approximately 300 time steps and are significantly lower compared to those of the pre-trained DPF. This indicates that the proposed OL-DPF can quickly adapt to new data when distribution shift occurs in this experiment setup. We report in Table~\ref{tab2} the mean and standard deviation of the overall RMSEs averaged across 5,000 time steps, confirming that the performance advantage of the OL-DPF compared with the pre-trained DPF. As expected, the RMSEs of the OL-DPF is slightly higher than those obtained by training DPFs with the oracle ground truth data in the test time.

\begin{table}[htbtp]
	\centering
	\caption{The comparison of test RMSEs in the online learning stage produced by different methods in the non-linear object tracking experiment. The reported mean and standard deviation are computed with 50 random simulations.}
	\label{tab2}  
	\begin{tabular}{|c|c|c|c|}
		
		\hline
		Method     & RMSE     \\ \hline
				Pre-trained DPF & $1110.84 \pm 341.67$   \\ \hline

		 OL-DPF    & $741.75 \pm 294.20$    \\ \hline
		
		DPF (oracle)    & $633.78 \pm 227.86$   \\ \hline
		
	\end{tabular}

\end{table}

\section{Conclusion}
This paper introduces an online learning framework for training differentiable particle filters with unlabelled data in an online manner. The proposed OL-DPFs address this by maximising an evidence lower bound on the conditional likelihood for each sliding window in the test-time trajectory. We empirically evaluated the performance of OL-DPFs on multivariate linear Gaussian state-space models with varying dimensionalities and a nonlinear position tracking experiment. Experimental results validated the proposed OL-DPFs' ability to adapt to new data distributions when distribution shift happens. We note, however, this work is our initial exploration in the direction of enabling DPFs to leverage unlabeled data for online learning.
There are a number of interesting further research directions, including a more theoretically justified and empirically effective loss function designed for online training of differentiable particle filters in more realistic simulation setups and real-world applications.

\label{sec:conclusion}
\vspace{12pt}

\bibliographystyle{IEEEtran}
\bibliography{ref} 

\begin{thebibliography}{10}
\providecommand{\url}[1]{#1}
\csname url@samestyle\endcsname
\providecommand{\newblock}{\relax}
\providecommand{\bibinfo}[2]{#2}
\providecommand{\BIBentrySTDinterwordspacing}{\spaceskip=0pt\relax}
\providecommand{\BIBentryALTinterwordstretchfactor}{4}
\providecommand{\BIBentryALTinterwordspacing}{\spaceskip=\fontdimen2\font plus
\BIBentryALTinterwordstretchfactor\fontdimen3\font minus \fontdimen4\font\relax}
\providecommand{\BIBforeignlanguage}[2]{{%
\expandafter\ifx\csname l@#1\endcsname\relax
\typeout{** WARNING: IEEEtran.bst: No hyphenation pattern has been}%
\typeout{** loaded for the language `#1'. Using the pattern for}%
\typeout{** the default language instead.}%
\else
\language=\csname l@#1\endcsname
\fi
#2}}
\providecommand{\BIBdecl}{\relax}
\BIBdecl

\bibitem{gordon1993novel}
N.~Gordon, D.~Salmond, and A.~Smith, ``{Novel approach to nonlinear/non-{Gaussian} {Bayesian} state estimation},'' in \emph{IEE Proc. F (Radar and Signal Process.)}, vol. 140, 1993, pp. 107--113.

\bibitem{djuric2003particle}
P.~M. Djuric, J.~H. Kotecha, J.~Zhang, Y.~Huang, T.~Ghirmai, M.~F. Bugallo, and J.~Miguez, ``Particle filtering,'' \emph{IEEE Signal Process. Mag.}, vol.~20, no.~5, pp. 19--38, 2003.

\bibitem{doucet2000sequential}
A.~Doucet, S.~Godsill, and C.~Andrieu, ``On sequential {Monte Carlo} sampling methods for {Bayesian} filtering,'' \emph{Stat. Comput}, vol.~10, no.~3, pp. 197--208, 2000.

\bibitem{qian20173d}
X.~Qian, A.~Brutti, M.~Omologo, and A.~Cavallaro, ``3d audio-visual speaker tracking with an adaptive particle filter,'' in \emph{Proc. IEEE Int. Conf. Acoust. Speech Signal Process (ICASSP)}, 2017, pp. 2896--2900.

\bibitem{creal2012survey}
D.~Creal, ``A survey of sequential {Monte Carlo} methods for economics and finance,'' \emph{Econometric Rev.}, vol.~31, no.~3, pp. 245--296, 2012.

\bibitem{pozna2022hybrid}
C.~Pozna, R.-E. Precup, E.~Horv{\'a}th, and E.~M. Petriu, ``Hybrid particle filter--particle swarm optimization algorithm and application to fuzzy controlled servo systems,'' \emph{IEEE Trans. Fuzzy Syst.}, vol.~30, no.~10, pp. 4286--4297, 2022.

\bibitem{kantas2009overview}
N.~Kantas, A.~Doucet, S.~S. Singh, and J.~M. Maciejowski, ``An overview of sequential {M}onte {C}arlo methods for parameter estimation in general state-space models,'' \emph{IFAC Proc. Vol.}, vol.~42, no.~10, pp. 774--785, 2009.

\bibitem{kantas2015particle}
N.~Kantas, A.~Doucet, S.~S. Singh, J.~Maciejowski, and N.~Chopin, ``On particle methods for parameter estimation in state-space models,'' \emph{Stat. Sci.}, 2015.

\bibitem{corenflos2021differentiable}
A.~Corenflos, J.~Thornton, G.~Deligiannidis, and A.~Doucet, ``Differentiable particle filtering via entropy-regularized optimal transport,'' in \emph{Proc. Int. Conf. Mach. Learn. (ICML)}, 2021, pp. 2100--2111.

\bibitem{karkus2018particle}
P.~Karkus, D.~Hsu, and W.~S. Lee, ``Particle filter networks with application to visual localization,'' in \emph{Proc. Conf. Robot. Learn. (CoRL)}, Zürich, Switzerland, 2018, pp. 169--178.

\bibitem{chen2023overview}
X.~Chen and Y.~Li, ``An overview of differentiable particle filters for data-adaptive sequential {Bayesian} inference,'' \emph{arXiv preprint arXiv:2302.09639}, 2023.

\bibitem{jonschkowski18}
R.~Jonschkowski, D.~Rastogi, and O.~Brock, ``Differentiable particle filters: end-to-end learning with algorithmic priors,'' in \emph{Proc. Robot.: Sci. and Syst. (RSS)}, Pittsburgh, Pennsylvania, July 2018.

\bibitem{chen2021differentiable}
X.~Chen, H.~Wen, and Y.~Li, ``Differentiable particle filters through conditional normalizing flow,'' in \emph{Proc. IEEE Int. Conf. Inf. Fusion. (FUSION)}, 2021, pp. 1--6.

\bibitem{gama2023unsupervised}
F.~Gama, N.~Zilberstein, M.~Sevilla, R.~Baraniuk, and S.~Segarra, ``Unsupervised learning of sampling distributions for particle filters,'' \emph{arXiv preprint arXiv:2302.01174}, 2023.

\bibitem{andrieu2010particle}
C.~Andrieu, A.~Doucet, and R.~Holenstein, ``Particle{ Markov chain Monte Carlo }methods,'' \emph{J. R. Stat. Soc. Ser. B. Stat. Methodol.}, vol.~72, no.~3, pp. 269--342, 2010.

\bibitem{chopin2013smc2}
N.~Chopin, P.~E. Jacob, and O.~Papaspiliopoulos, ``{SMC2}: an efficient algorithm for sequential analysis of state space models,'' \emph{J. R. Stat. Soc. Ser. B. Stat. Methodol.}, vol.~75, no.~3, pp. 397--426, 2013.

\bibitem{crisan2018nested}
D.~Crisan and J.~M{\'I}guez, ``Nested particle filters for online parameter estimation in discrete-time state-space {Markov} models,'' \emph{Bernoulli}, vol.~24, no.~4A, pp. 3039--3086, 2018.

\bibitem{chen2022conditional}
X.~Chen and Y.~Li, ``Conditional measurement density estimation in sequential {Monte Carlo} via normalizing flow,'' in \emph{Proc. Euro. Sig. Process. Conf. (EUSIPCO)}, 2022, pp. 782--786.

\bibitem{ionides2006inference}
E.~L. Ionides, C.~Bret{\'o}, and A.~A. King, ``Inference for nonlinear dynamical systems,'' \emph{Proc. Natl. Acad. Sci.}, vol. 103, no.~49, pp. 18\,438--18\,443, 2006.

\bibitem{ionides2015inference}
E.~L. Ionides, D.~Nguyen, Y.~Atchad{\'e}, S.~Stoev, and A.~A. King, ``Inference for dynamic and latent variable models via iterated, perturbed bayes maps,'' \emph{Proc. Natl. Acad. Sci.}, vol. 112, no.~3, pp. 719--724, 2015.

\bibitem{douc2005comparison}
R.~Douc and O.~Capp{\'e}, ``Comparison of resampling schemes for particle filtering,'' in \emph{Proc. Int. Symp. Image and Signal Process. and Anal.}, Zagreb, Croatia, 2005.

\bibitem{le2018auto}
T.~A. Le, M.~Igl, T.~Rainforth, T.~Jin, and F.~Wood, ``Auto-encoding sequential {Monte Carlo},'' in \emph{Proc. Int. Conf. Learn. Rep. (ICLR)}, Vancouver, Canada, Apr. 2018.

\bibitem{maddison2017filtering}
C.~J. Maddison, J.~Lawson, G.~Tucker, N.~Heess, M.~Norouzi, A.~Mnih, A.~Doucet, and Y.~Teh, ``Filtering variational objectives,'' \emph{in Proc. Adv. Neural Inf. Process. Syst. (NeurIPS)}, vol.~30, 2017.

\bibitem{naesseth2018variational}
C.~Naesseth, S.~Linderman, R.~Ranganath, and D.~Blei, ``Variational sequential {Monte Carlo},'' in \emph{Proc. Int. Conf. Artif. Intel. and Stat. (AISTATS)}, Playa Blanca, Spain, Apr. 2018.

\bibitem{chopin2020particle}
N.~Chopin, O.~Papaspiliopoulos, N.~Chopin, and O.~Papaspiliopoulos, ``Particle filtering,'' \emph{An Introduction to Sequential Monte Carlo}, pp. 129--165, 2020.

\bibitem{bugallo2007performance}
M.~F. Bugallo, S.~Xu, and P.~M. Djuri{\'c}, ``Performance comparison of ekf and particle filtering methods for maneuvering targets,'' \emph{Digit. Signal Process.}, vol.~17, no.~4, pp. 774--786, 2007.

\bibitem{kingma2015adam}
D.~P. Kingma and J.~Ba, ``Adam: {A} method for stochastic optimization,'' in \emph{Proc. Int. Conf. on Learn. Represent. (ICLR)}, San Diego, USA, May 2015.

\end{thebibliography}
\end{document}